\pdfoutput=1

\documentclass[11pt]{article}

\usepackage[final]{EMNLP2022}

\usepackage{times}
\usepackage{latexsym}

\usepackage[T1]{fontenc}

\usepackage[utf8]{inputenc}

\usepackage{microtype}
\usepackage{booktabs}
\usepackage{multirow}
\usepackage{geometry}
\usepackage{CJKutf8}
\usepackage{graphicx}
\usepackage{color}
\usepackage{bm}
\usepackage{bbding}
\usepackage{array}
\usepackage{subfigure}
\usepackage{amsmath}
\usepackage{footmisc}
\usepackage{microtype}
\usepackage{makecell}
\usepackage{amsmath,amssymb}
\usepackage{color}
\usepackage[normalem]{ulem}
\useunder{\uline}{\ul}{}

\usepackage{inconsolata}

\title{Learning Instructions with Unlabeled Data for Zero-Shot Cross-Task Generalization}

\author{
  Yuxian Gu, 
  Pei Ke,
  Xiaoyan Zhu,
  Minlie Huang$^\dagger$\\
  The CoAI group, Tsinghua University, Beijing, China \\
  Institute for Artificial Intelligence, State Key Lab of Intelligent Technology and Systems, \\
Beijing National Research Center for Information Science and Technology, \\
Department of Computer Science and Technology, Tsinghua University, Beijing, China \\
  \texttt{guyx21@mails.tsinghua.edu.cn}, \ \ \ \texttt{kepei1106@outlook.com}\\
  \texttt{\{zxy-dcs,aihuang\}@tsinghua.edu.cn}\\
 }

\begin{document}
\maketitle
\begin{abstract}
Training language models to learn from human instructions for zero-shot cross-task generalization has attracted much attention in NLP communities. Recently, instruction tuning (IT), which fine-tunes a pre-trained language model on a massive collection of tasks described via human-craft instructions, has been shown effective in instruction learning for unseen tasks. However, IT relies on a large amount of human-annotated samples, which restricts its generalization. Unlike labeled data, unlabeled data are often massive and cheap to obtain. In this work, we study how IT can be improved with unlabeled data. We first empirically explore the IT performance trends versus the number of labeled data, instructions, and training tasks. We find it critical to enlarge the number of training instructions, and the instructions can be underutilized due to the scarcity of labeled data. Then, we propose \textbf{U}nlabeled \textbf{D}ata Augmented \textbf{I}nstruction \textbf{T}uning (UDIT) to take better advantage of the instructions during IT by constructing pseudo-labeled data from unlabeled plain texts. We conduct extensive experiments to show UDIT's effectiveness in various scenarios of tasks and datasets. We also comprehensively analyze the key factors of UDIT to investigate how to better improve IT with unlabeled data. The code is publicly available at \url{https://github.com/thu-coai/UDIT}. 

\end{abstract}

\section{Introduction}

{\let\thefootnote\relax\footnotetext{
$^\dagger$ Corresponding author. }
}

\begin{figure}[t]
    \centering
    \includegraphics[width=\linewidth]{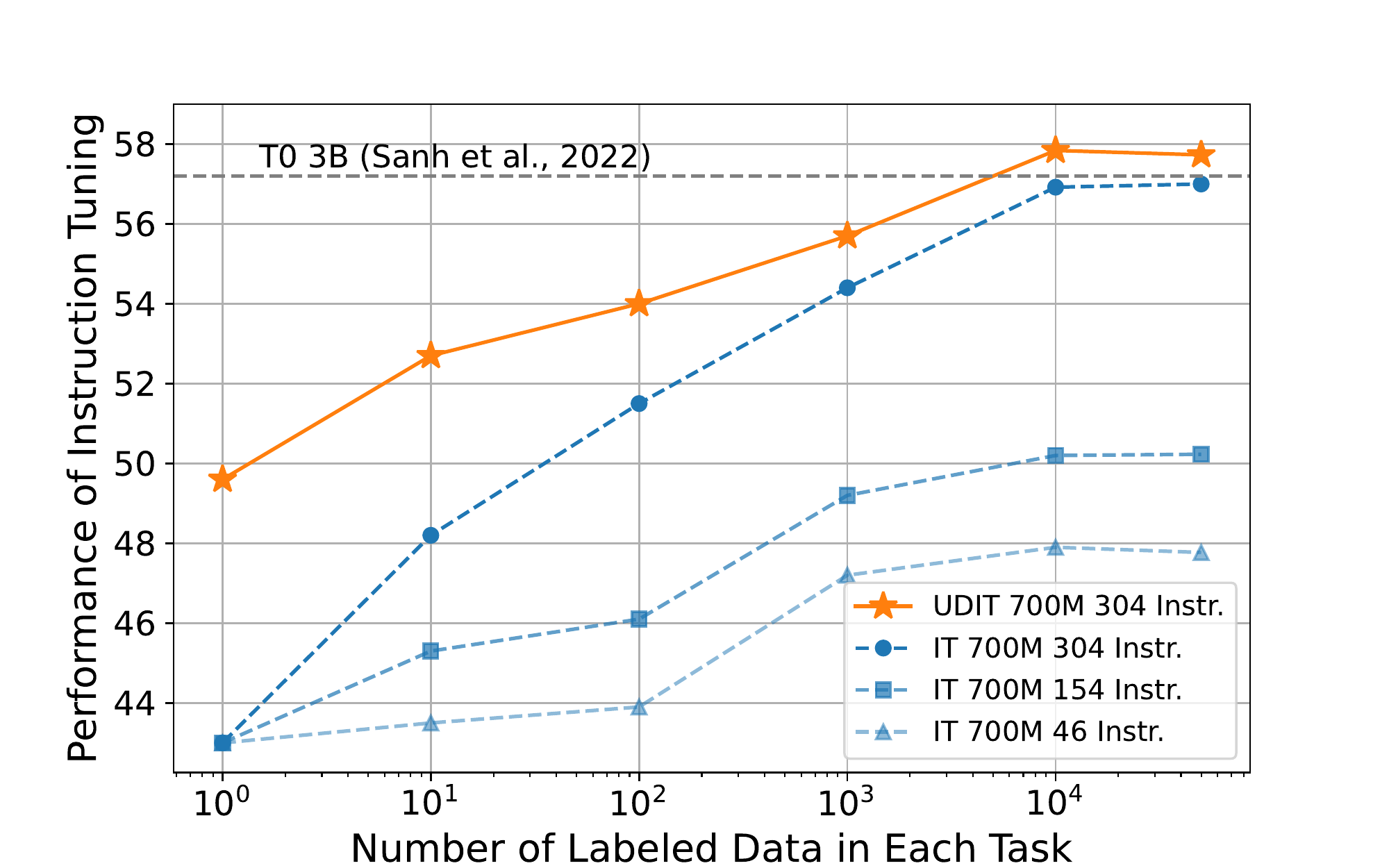}
    \caption{The performance of IT and UDIT with respect to the instruction numbers and labeled data amounts. We follow \citet{t0} to fine-tune a 700M PLM and then test its zero-shot generalization ability on unseen tasks. The x-axis represents the number of labeled samples in each training task, and the y-axis is the average performance on evaluation tasks. We control the instruction numbers by gradually adding training tasks.}
    \label{fig:setting}
\end{figure}

The instruction learning paradigm~\cite{zest}, where language models learn from human instructions to perform unseen tasks in zero-shot scenarios, has received increasing attention recently. Compared to conventional machine learning paradigms that mainly learn from data examples, instruction learning requires models to complete tasks based on the understanding of human-written task descriptions without task-specific data, which is closer to general AI systems. For instance, in summarization tasks, a model is only given an explicit instruction ``Summarize the following article in brief:'' and an article to generate the corresponding summary. To realize instruction learning, recent works such as FLAN~\cite{flan} and T0~\cite{t0} propose instruction tuning (IT), which fine-tunes pre-trained language models (PLMs)~\cite{plmsurvey} on a large collection of tasks with human-annotated data specified in descriptive instructions. Through IT, PLMs learn to follow the human-written instructions to complete the corresponding tasks, which enables them to perform instruction learning in unseen tasks.

An intuitive way to boost the performance of IT is to increase the number of training instructions and data examples. As shown in Figure \ref{fig:setting}, the number of training instructions largely determines the best performance of IT, and the corresponding human-annotated data should be also sufficient for the model to learn these instructions well. However, the amount and domain diversity of labeled data in different tasks vary greatly. In practice, many low-resource tasks lack sufficient multi-domain human-annotated examples. This can lead to easy overfitting to specific domains or examples when learning the corresponding instructions, which affects the zero-shot performance in instruction learning. 

Introducing unlabeled data is a common approach to alleviating the data scarcity problem in supervised learning~\cite{gpt3, uda, self-training-pre-training} because large-scale unlabeled plain texts are much easier to access. However, we argue that their benefit to IT is still inconclusive. This is because IT is much more challenging, requiring learning the mapping between human instructions and task semantics rather than that between samples and labels in a single task.


Therefore, in this work, we investigate incorporating unlabeled data into instruction learning. We focus on the two questions: (1) \textit{Is it possible to perform IT with unlabeled plain texts when there are few or even no human-annotated data?} and (2) \textit{How to better use unlabeled plain texts to improve IT for zero-shot cross-task generalization?} 

To study (1), we propose \textbf{U}nlabeled \textbf{D}ata Augmented \textbf{I}nstruction \textbf{T}uning (UDIT) to effectively use unlabeled data to help instruction learning. Specifically, we construct pseudo-labeled data from unlabeled plain texts according to task instructions.
The pseudo-labeled data which enlarge training samples and diversify data domains help to learn the meanings of the corresponding task instructions better.
We test UDIT under various scenarios of training tasks and labeled data to verify that learning instructions from unlabeled data is possible.

To study (2), we compare UDIT with previous methods 
to show its superior performance in using unlabeled data. We also conduct extensive experiments to reveal the underlying factors to the success of UDIT. Specifically, our contributions are summarized as follows:
\begin{itemize}
    \item We introduce UDIT, a training framework that incorporates unlabeled data into instruction tuning for zero-shot cross-task generalization.
    \item Through UDIT, we empirically verify that PLMs can learn to follow human-written instructions with unlabeled data when there are few or even no annotated samples.
    \item We show that UDIT is a significantly better way to use unlabeled data to improve instruction tuning, making a 700M PLM with UDIT outperform the 3B counterpart based on IT.
    \item We comprehensively analyze the key factors of UDIT and give some insights into using unlabeled data to improve instruction learning.
\end{itemize}

\begin{figure*}[t]
    \centering
    \includegraphics[width=\linewidth]{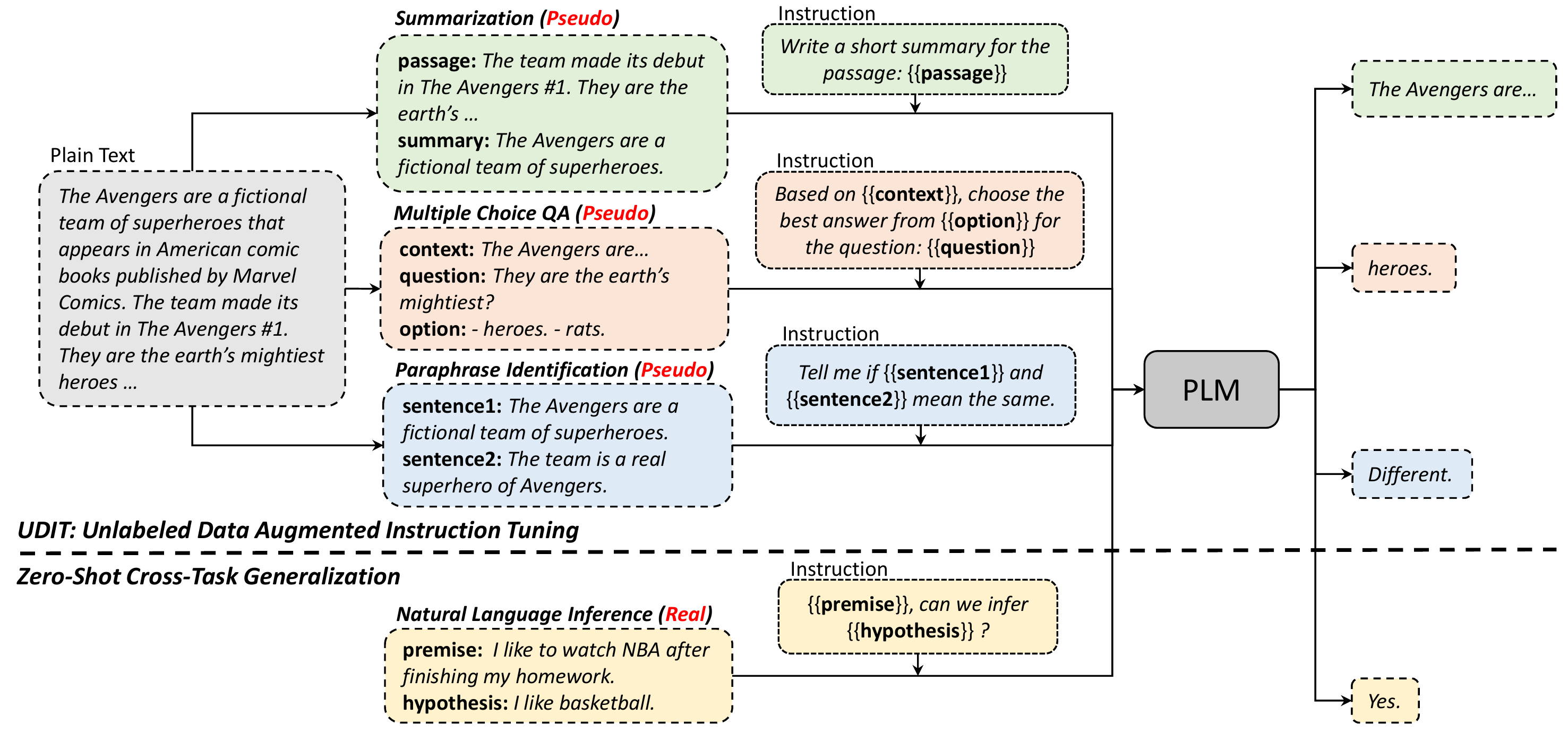}
    \caption{An Example of UDIT. The pseudo-labeled data of Summarization, Multiple-Choice QA, and Paraphrase Identification are constructed from plain texts. Then the corresponding instructions are applied to these samples. We fine-tune the PLM on these samples in a multi-task text-to-text language modeling objective and test it for zero-shot cross-task generalization where the evaluation task (Natural Language Inference) is unseen during training.}
    \label{fig:method}
    \vspace{-0.5em}
\end{figure*}

\section{Related Works}
\paragraph{Instruction Learning.}
Recently, large PLMs like GPT-3~\cite{gpt3} have shown promising performance in learning from human instructions to solve tasks in few-shot and zero-shot scenarios~\cite{prompt_survey}. Several works propose benchmarks~\cite{zest,turking-test,natural-instructions,regset} to evaluate instruction learning for zero-shot cross-task generalization~\cite{crossfit}. To enhance instruction understanding, many works adopt IT, which fine-tunes PLMs on massive task clusters described by instructions in a multi-task fashion, such as FLAN~\cite{flan}, T0~\cite{t0}, ZeroPrompt~\cite{zero-prompt}, and InstructGPT~\cite{instruct-gpt}. These models show superior zero-shot performance on unseen tasks. To better understand IT and zero-shot learning of PLMs, \citet{it-exp} compares different model architectures and training objectives. Some works also incorporate unlabeled data to improve the zero-shot performance of IT~\cite{prompt-consistency, recross}. But they assume the existence of unlabeled samples in evaluation tasks, while we only use plain texts, which is more in line with the zero-shot cross-task evaluation scenario.

\paragraph{Semi-Supervised Learning.} Semi-supervised learning adopts unlabeled data to improve supervised learners~\cite{semi-supervised-learning}. Many previous works use consistency training to regularize model predictions~\cite{bachman2014learning, rasmus2015semi, uda}. Self-training ~\cite{self-training-first,self-training-pre-training} is also widely used, which assigns synthetic labels to unlabeled data with a teacher model. These data are then used to train the student model. However, these methods typically assume the availability of unannotated task-specific data while we focus on using task-agnostic plain texts, which is more practical.


\paragraph{Self-Supervised Training in NLP.} Training with self-supervised tasks is also related to our method, which helps models obtain versatile knowledge from large-scale plain texts and boosts the model performance~\cite{bert,gpt2,t5,bart,albert,cert}. Some works also find that carefully designed self-supervised tasks can bring further improvement to low-resource tasks~\cite{self_sup_meta, ppt,self-sup}. However, conventional self-supervised tasks are designed independent of human instructions~\cite{loss_modern_lm}, while tasks in UDIT match the instruction semantics closely, which is crucial to instruction learning for zero-shot cross-task generalization.

\section{Method}
\subsection{Background}
In this section, we first give a formal description of IT. We define a ``task'' as a pair $\left(D, I\right)$, where $D$ is the task-specific dataset, and $I$ is a set of instructions describing the task.
We assume that the tasks can be divided into $n$ clusters $\mathcal{T} = \{T_1, T_2, \cdots, T_n\}$ according to the task similarities, where  the $i^{\text{th}}$ cluster $T_i=\{ (D_1, I_1), (D_2, I_2), \cdots, (D_{k_i}, I_{k_i}) \}$ contains $k_i$ tasks.
For example, in the cluster ``Multiple-Choice QA'', a data sample typically consists of a passage, a question, several answer options, and the answer. As shown in Figure \ref{fig:method}, the instructions serve as templates to convert the inputs and outputs to natural texts and formulate all tasks into text-to-text language modeling problems. In IT, a PLM is first fine-tuned on several clusters $\mathcal{T}_{\text{Train}} \subsetneqq \mathcal{T}$ in a multi-task fashion. Then the model is evaluated on the tasks in novel clusters $\mathcal{T}_{\text{Test}} = \mathcal{T} \backslash \mathcal{T}_{\text{Train}}$ with instructions only, as shown in the ``Zero-Shot Cross-Task Generalization'' part of Figure \ref{fig:method}. 

In this paper, we mainly follow the settings of T0~\cite{t0}, including the training tasks, instructions, and the split of task clusters. However, our findings can also be applied to other scenarios. T0 is a representative model instruction-tuned on 8 task clusters based on the pre-trained T5 model~\cite{t5} and tested on 4 task clusters. The instructions are collected from the Public Pool of Prompts (P3)~\cite{p3} which contains thousands of crowdsourced instructions.

\subsection{Overview}
\label{sec:overview}

Figure \ref{fig:method} shows an overview of UDIT. To better learn the instructions in $\mathcal{T}_\text{Train}$, we construct pseudo-labeled data from the unlabeled plain texts according to the meaning of the instructions in each task cluster $T_i\in \mathcal{T}_{\text{Train}}$. The plain texts are a mixture of multi-domain corpora, including BookCorpus \cite{bookcorpus}, CC-News \cite{cc_news}, OpenWebText \cite{openwebtext}, Wikipedia \cite{wikidump}, and IMDB Review \cite{imdb}, totaling about 37.2G. The details of these corpora are shown in Appendix \ref{app:data_info}. The constructing process is based on heuristic rules, widely used NLP toolkits like NLTK\footnote{\url{https://www.nltk.org/}}, and basic data augmentation techniques like back-translation~\cite{back-translation}. Then, we apply the instructions in $T_i$ to the pseudo-labeled samples and fine-tune the PLM on pseudo-labeled and labeled samples with a multi-task language modeling objective. Although the pseudo-labeled data are constructed at the level of task clusters rather than single tasks, we find they match the meanings of most instructions in the corresponding cluster due to the task similarities. Note that we do not assume the existence of labeled data during the constructing process, which means that UDIT is applicable under various settings with or without labeled data. The following section briefly introduces the pseudo-labeled data construction for the 8 task clusters in T0. We provide some examples of the constructed data in Appendix \ref{app:case}.


\subsection{Constructing Pseudo-Labeled Data}
\paragraph{Multiple-Choice QA (MCQA).} The sample in MCQA consists of a passage, a related question, an answer, and several options. Given a plain-text document, we design two methods to construct pseudo-labeled data: (1) We first randomly replace one noun in a randomly selected sentence with a "\_" symbol. Then, we add a ``?'' mark to the end of the sentence to form a question. We treat the texts before the sentence as the passage and the replaced word as the answer. The options are sampled from the words with the same part of speech as the answer. (2) We observe that many questions are naturally followed by its answer in our corpus. Therefore, we search for questions in the document and treat previous texts as the passage and the following sentence as the answer. The options are sampled from the sentences after the answer.

\paragraph{Extractive QA (EXQA).} EXQA aims to answer the questions using the phrases in the given passages. We mainly follow \citet{uqa} which first selects entities in the plain-text documents as the answers and uses templates to convert the sentences containing the answers to questions.

\paragraph{Close-Book QA (CBQA).} CBQA is similar to EXQA except for the absence of the passage. Therefore, we use the question-answer pair from the pseudo-labeled data of EXQA.

\paragraph{Sentiment (SENT).} SENT requires identifying the sentiment labels of given texts. We use a keyword-based sentiment analyzer in NLTK to annotate sentiment labels. To improve the label quality, we only construct pseudo-labeled data from the IMDB Review corpus for this cluster.

\paragraph{Topic Classification (TC).} TC requires finding proper topic labels for input passages. We notice that many URLs of the CC-News corpus contain the topic of passages. Therefore, we devise heuristic rules to extract topic labels from the URLs to build the pseudo-labeled data. We first split the URL by ``/'' and search for topic words from left to right. We stop at the first string that is composed of English letters, shorter than 20 characters, and not in [``news'', ``en'', ``story'', ``us'', ``articles'', ``local'', ``english'', ``tag'', ``post'']. Then, we choose the most frequent 14 strings as the topic labels and the corresponding passages as the inputs.

\paragraph{Structure-to-Text (S2T).} S2T requires generating natural sentences that describe input structural data like graphs. Since the input data are usually linearized as word sequences in the instructions, we adopt a keyword-to-text generation task that takes a random subset of the notional words in a sentence as the input and the sentence as the output.

\paragraph{Summarization (SUM).} For summarization, we adopt Leading Sentence Generation (LSG) and Gap Sentence Generation (GSG) from~\citet{psp}. LSG takes the title of a passage as the summary and the body as the input. GSG treats the sentence overlapping other document parts the most as the summary and the remaining sentences as the input.

\paragraph{Paraphrase Identification (PARA).} PARA aims to identify whether two sentences have the same meaning. Given a sentence $s$ in the plain texts, we add word-level perturbation to $s$ to get $s_{\text{pert}}$. We consider two kinds of perturbations: (1) Randomly choosing a word and replacing it with its antonym via NLTK. (2) Picking out nouns in the sentence via NLTK and shuffling their order. Then we get $\widetilde{s}$ and $\widetilde{s}_{\text{pert}}$ by adopting back-translation to $s$ and $s_{\text{pert}}$, respectively. We treat $(s,\widetilde{s})$ as the positive pair and $(s, \widetilde{s}_{\text{pert}})$ as the negative pair.

\section{Experiment}
\subsection{Setup}
\label{sec:exp_setup}
\paragraph{Settings.} We consider three scenarios in which unlabeled data can be utilized to enhance instruction learning: (1) \textbf{No Labeled Data}, where only the instructions for each task are available. (2) \textbf{Few Labeled Data}, where only a small part of the labeled data is available. (3) \textbf{Full Labeled Data}, where all the labeled data are available during IT. 

\begin{table*}[t]
\centering
\small
\begin{tabular}{lrcc|ccc|ccc|c|c}
\toprule
\multirow{2}{*}{Method} & \multirow{2}{*}{Size} & \multicolumn{2}{c|}{Coref.} & \multicolumn{3}{c|}{NLI} & \multicolumn{3}{c|}{Sentence Comp.} & WSD  & \multirow{2}{*}{Avg.} \\ \cmidrule(l){3-11}
                        &                       & WSC          & Wino.        & CB     & RTE    & ANLI   & COPA     & H-Swag     & Story    & WiC  &                       \\ \midrule
DirectZS                & 700M                  & 50.3         & 50.9         & 32.8   & 48.4   & 32.8   & 42.3     & 26.4       & 52.7     & 50.9 & 43.0                  \\
DirectZS                & 3B                    & 49.5         & 50.9         & 31.1   & 47.9   & 32.5   & 46.4     & 25.7       & 55.3     & 50.5 & 43.3                  \\
DirectZS                & 11B                   & \textbf{54.1}& 50.6         & 34.3   & 53.0   & \underline{32.9} & 54.9     & 27.0       & 48.2     & 50.3 & 45.0                  \\ \midrule
ExtraLM                 & 700M                  & 52.0         & 51.5         & 29.3   & 52.3   & 32.3   & 48.7     & 23.9       & 51.6     & 50.5 & 43.6                  \\
SelfSup-IT              & 700M                  & 50.5          &\textbf{54.0} & 41.9   & 53.0   & \underline{32.9} & 50.7     & 24.0       & 51.4     & 50.2 & 45.4                  \\ \midrule
UDIT                    & 700M              &\underline{53.6} & 52.9    & \underline{44.2} & \underline{54.0} & \textbf{33.0} & \underline{59.4} & \textbf{29.4}       & \underline{67.4} & \textbf{53.0} & \underline{49.6}   \\
UDIT + SelfSup-IT      & 700M               & \underline{53.6} & \underline{53.1} & \textbf{45.1}   & \textbf{56.6} & 32.4   &  \textbf{59.5} & \underline{28.1} & \textbf{71.8} & \underline{52.0} & \textbf{50.2}                      \\ \bottomrule
\end{tabular}
\caption{The zero-shot cross-task generalization results of classification tasks in the ``No Labeled Data'' scenario. We report the average accuracy of different testing instructions on the official validation set of each dataset. We reprint the DirectZS scores of the 11B model from \citet{t0}. The best results on each dataset are in \textbf{boldface} and the second-best results are \underline{underlined}.}
\label{tab:exp_none}
\vspace{-0.5em}
\end{table*}

\paragraph{Datasets.} Following \citet{t0}, we use 8 task clusters as $\mathcal{T}_\text{Train}$, which contains 36 datasets and 304 instructions. $\mathcal{T}_\text{Test}$ contains 6 task clusters consisting of 9 text classification tasks\footnote{The ANLI task contains datasets of 3 versions that share the same instructions. We only evaluate our model on the R1 version for simplicity.} and 2 language generation tasks. Detailed data information can be found in Appendix \ref{app:data_info}.

\paragraph{Training and Evaluation Details.} For computational efficiency, we conduct our experiments mainly based on a 700M T5 model. We mix the labeled and pseudo-labeled data for multi-task fine-tuning. Unless specified otherwise, we use at most 10k labeled/pseudo-labeled samples for each task because we find more samples bring little improvement. We choose the best checkpoint on the merged validation splits of datasets in $\mathcal{T}_{\text{Train}}$ for evaluation. More hyper-parameter details are shown in Appendix \ref{app:hyper_param}. In evaluation, we report the mean (Section \ref{sec:res}) and median (Appendix \ref{app:res_median}) of the performance across different instructions on the validation set of each task in $\mathcal{T}_\text{Test}$. For the multiple-choice tasks, we select the option with the highest log-likelihood~\cite{gpt3} as the answer.

\paragraph{Baselines.} We consider the following baselines:

    \noindent (1) \textit{Direct Zero-Shot} (\textbf{DirectZS}): The PLM is directly evaluated on $\mathcal{T}_\text{Test}$ without fine-tuning.
    
    \noindent (2) \textit{Vanilla Instruction Tuning} (\textbf{Vanilla-IT}): The model is instruction-tuned on the labeled data in $\mathcal{T}_\text{Train}$, which stays the same with \citet{t0}.
    
    \noindent (3) \textit{Self-Supervised Training}: Besides the labeled data in IT, the model is also tuned on our unlabeled plain-text corpus with the language modeling objective (\textbf{ExtraLM}) or the four self-supervised objectives proposed in \citet{self-sup} (\textbf{SelfSup-IT}). The proportion of training samples to our pseudo-labeled samples is 1:1.
    
    \noindent (4) \textit{Data Augmentation} (\textbf{DataAug-IT}): For the tasks with few labeled data, we perform back-translation and augment the labeled data to twice as large~\cite{back-translation-new}. 

\subsection{Results}
\label{sec:res}
\subsubsection{No Labeled Data}
\label{sec:no_label_data}

Table \ref{tab:exp_none} shows the results where no labeled data are available, from which we have 3 observations. 

\textit{First}, all methods that use unlabeled data (ExtraLM, SelfSup-IT, and UDIT) outperform DirectZS, suggesting that PLMs can learn to follow instructions for zero-shot cross-task generalization with unlabeled data when human-labeled samples are absent. 

\textit{Second}, among different methods using unlabeled data, self-supervised training only brings marginal improvement, while UDIT boosts the performance largely on most tasks. This indicates that using unlabeled data to improve instruction learning is non-trivial. Simple self-supervised tasks cannot reflect the characteristics of human instructions, while UDIT directly helps the PLM learn the mapping between instructions and task semantics.

\textit{Third}, UDIT can be combined with self-supervised training when we mix the training samples augmented by these two methods.
The row ``UDIT + SelfSup-IT'' achieves the best average performance, which means that these two methods are complementary in this scenario.

\subsubsection{Few Labeled Data}
\label{sec:little_labeled_data}
\begin{table*}[t]
\setlength\tabcolsep{5.8pt}
\centering
\small
\begin{tabular}{llcc|ccc|ccc|c|c}
\toprule
 \multirow{2}{*}{Setting} & \multirow{2}{*}{Method} & \multicolumn{2}{c|}{Coref.} & \multicolumn{3}{c|}{NLI} & \multicolumn{3}{c|}{Sentence Comp.}   & WSD  & \multirow{2}{*}{Avg.} \\ \cmidrule(l){3-11}
                          &                         & WSC          & Wino.        & CB     & RTE    & ANLI   & COPA     & H-Swag             & Story & WiC  &                       \\ \midrule
\multirow{4}{1.1cm}{Few\\Tasks}
                          & Vannila-IT              & 51.4         & 53.0         & 50.3   & 53.7   & 31.5   & \textbf{60.9} & \textbf{28.3} & 51.6  & 50.3 & 47.9                  \\
                          & SelfSup-IT              & 50.6         & 55.2         & 50.4   & 53.2   & \textbf{33.1} & 57.2 & 27.7            & 52.9  & 50.4 & 47.8                  \\ \cmidrule(lr){2-12}
                          & UDIT                    & \textbf{54.4}& 54.1         & \textbf{57.5} & \textbf{57.2} & 32.2 & \textbf{60.9} & 27.6 & 66.2 & 52.0 & \textbf{51.3}            \\
                          & UDIT + SelfSup-IT        & 51.7         & \textbf{55.8} & 48.0  & 56.2   & 32.9   & 60.4     & 26.3      & \textbf{70.0} & \textbf{52.1} & 50.4          \\ \midrule \midrule
\multirow{4}{1.1cm}{Few\\Datasets}
                          & Vannila-IT              & 49.8         & 51.3         & 32.8   & 50.5   & \textbf{32.9}   & 59.0     & 25.8       & 53.1     & 51.0 & 45.1                  \\
                          & SelfSup-IT              & 50.6         & 54.1         & 41.9   & 54.0   & \textbf{32.9}   & 55.0     & 26.4       & 59.0     & 51.4 & 47.2                   \\ \cmidrule(lr){2-12}
                          & UDIT                    & \textbf{51.1} & 51.2        & 48.3   & \textbf{65.9} & 31.8 & \textbf{66.8} & \textbf{29.4} & 71.0 & 51.6 & \textbf{51.9}            \\
                          & UDIT + SelfSup-IT        & 50.8         & \textbf{55.1} & \textbf{48.8} & 57.9 & 32.0 & 60.5 & 27.4      & \textbf{75.9}  & \textbf{53.2} & 51.3                    \\ \midrule \midrule
\multirow{6}{1.1cm}{Few\\Samples}
                          & Vanilla-IT              & 52.0         & 51.6         & 54.2   & 58.2   & 31.2   & 65.9     & 26.9              & 71.5     & 51.7 & 51.5                  \\
                          & SelfSup-IT              & 50.8         & 53.3         & 47.2   & 63.8   & 31.7   & 60.9     & 26.5              & 73.1     & 54.0 & 51.2                  \\
                          & DataAug-IT              & 50.9         & 52.0     & \textbf{55.1}& 56.3 & 31.2   & 67.2     & 28.1              & 72.0     & 53.0 & 51.8                  \\ \cmidrule(lr){2-12}
                          & UDIT                    & \textbf{53.8} & 52.9        & 51.0   & \textbf{64.4} & 31.3 & 68.4 & \textbf{30.5}     & 80.0  & 54.2 & \textbf{54.0}            \\
                          & UDIT + SelfSup-IT        & 51.6         & \textbf{54.3}& 51.8   & 59.9   & \textbf{32.4} & 68.6 & 27.9    & \textbf{81.0} & \textbf{55.4} & 53.6                     \\
                          & UDIT + DataAug-IT        & 53.4         & 51.6         & 48.5   & 61.3   & 31.5   & \textbf{70.7} & 29.7         & 79.4   & 53.0 & 53.2                      \\ \bottomrule
\end{tabular}
\caption{The scores of classification accuracy in the ``Few Labeled Data'' scenario. In the ``Few Tasks'' block, IT is performed on EXQA, and UDIT adds pseudo-labeled data to other clusters. In the ``Few Datasets'' block, IT is performed on 10\% tasks from each cluster, and UDIT adds pseudo-labeled data to the remaining tasks. In the ``Few Samples'' block, IT is performed on 100 samples from each dataset, and UDIT adds pseudo-labeled data to all tasks. All experiments are based on the 700M model. The best scores in each block are in \textbf{boldface}.}
\label{tab:exp_little}
\vspace{-0.5em}
\end{table*}

In this section, we study a more practical scenario where only a small set of labeled data are available for IT and show the results in Table \ref{tab:exp_little}. We explore three different data scarcity settings. \underline{``Few Tasks''} simulates the setting where only a few task clusters have enough labeled data. Here, we choose EXQA as the data-sufficient cluster, where each task contains 10K samples. The results of other choices are shown in Appendix \ref{app:few_task}. \underline{``Few Datasets''} means only 10\% human-labeled datasets exist in each task cluster. And the \underline{``Few Samples''} block shows the results where IT is performed on all task clusters, but each dataset contains only 100 samples. Note that data augmentation (DataAug-IT) can only be applied to the ``Few Samples'' setting because there are no source data for back-translation in other settings. UDIT adds the pseudo-labeled data to those data-scarce tasks to enhance instruction learning.

Our findings from Table \ref{tab:exp_little} are as follows: 
\begin{enumerate}
    \vspace{-0.1cm}
    \item SelfSup-IT and DataAug-IT fail to bring significant improvement over Vanilla-IT. It is probably because the self-supervised tasks do not use instructions, and the augmented data are too similar to the source samples.
    \vspace{-0.1cm}
    \item UDIT performs the best on average under all the three settings, indicating that learning instruction semantics and training on sufficient diverse data are crucial to zero-shot cross-task generalization of PLMs.
    \vspace{-0.1cm}
    \item Unlike the observations in Section \ref{sec:no_label_data}, the benefit of combining self-supervised tasks with UDIT vanishes with the existence of the few labeled data.
\end{enumerate}

\begin{figure}[t]
    \subfigbottomskip=0cm
    \subfigure[Task Clusters] { \label{fig:task_cluster} 
    \centering
    \includegraphics[width=0.58\linewidth]{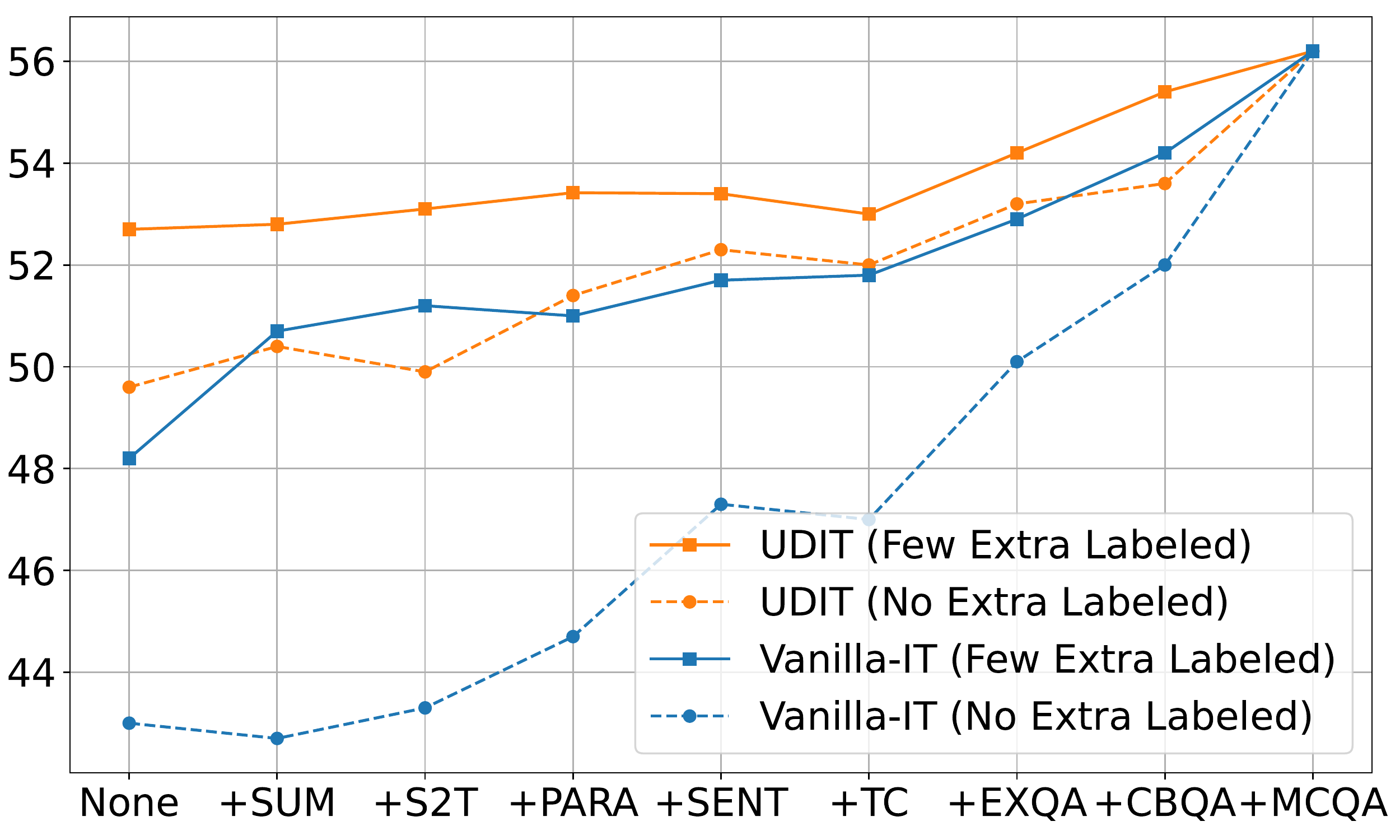} 
    }
    \subfigure[Task Proportions] { \label{fig:task_proportion} 
    \centering
    \includegraphics[width=0.36\linewidth]{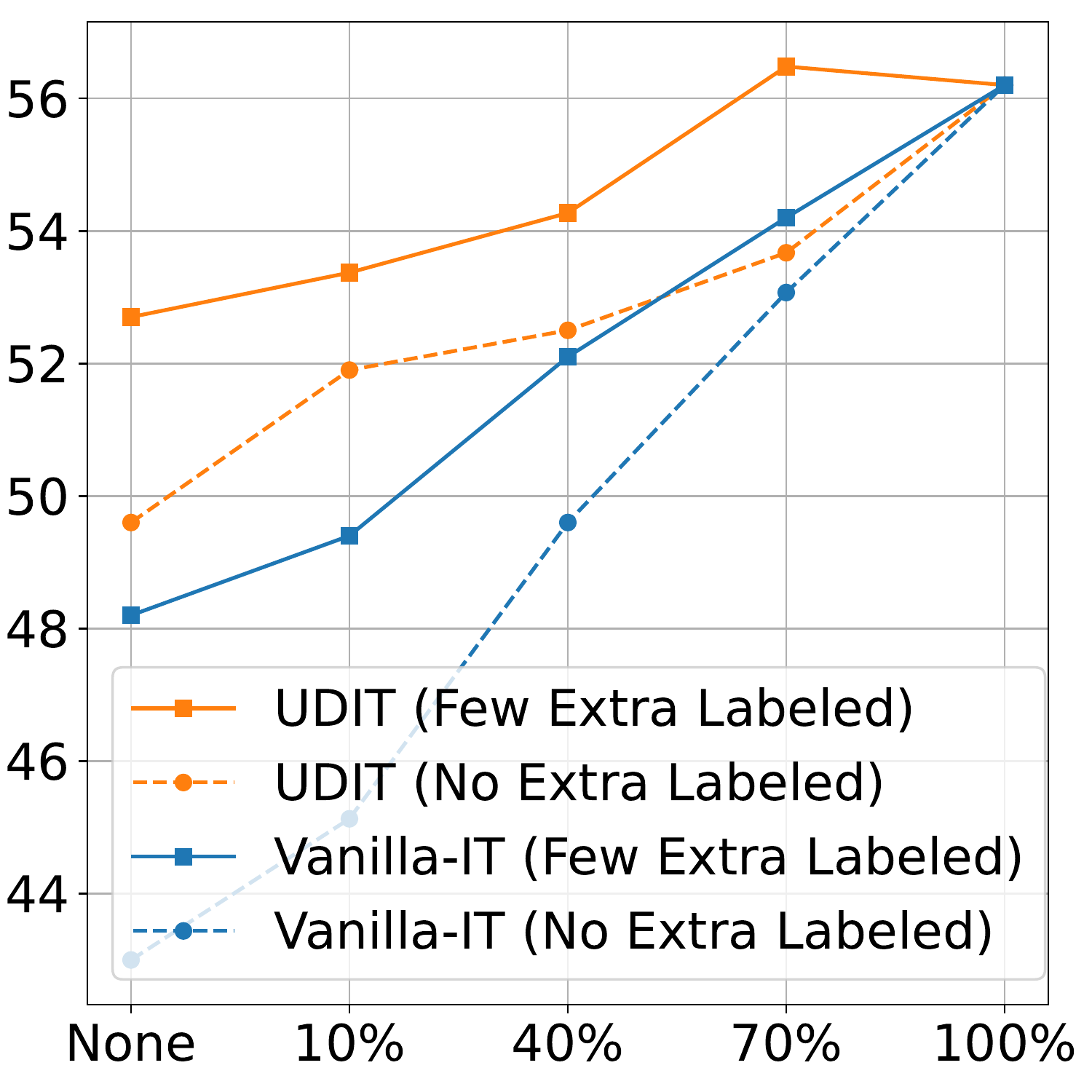} 
    } 
    \caption{The performance trend when number of data-sufficient tasks varies. The y-axis means the average classification results. ``Few/No Extra Labeled'' means there are few/no labeled data in the data-scarce tasks.} 
    \vspace{-1em}
    \label{fig:few_task} 
\end{figure} 

We also evaluate Vanilla-IT and UDIT when the number of data-sufficient tasks varies. In Figure \ref{fig:task_cluster}, we incrementally add task clusters containing full labeled data. And in Figure \ref{fig:task_proportion}, we gradually increase the proportion of data-sufficient tasks in each cluster. In these processes, other tasks are considered data-scarce, containing 100 (Few Extra Labeled) or no (No Extra Labeled) labeled samples to simulate the situation when both data-sufficient and data-scarce tasks exist. UDIT enhances the data-scarce tasks with pseudo-labeled data.\footnote{We take ``+TC'' in Figure \ref{fig:task_cluster} as an example. All tasks in SUM, S2T, PARA, SENT, and TC have enough data (10K samples). Each task in EXQA, CBQA, and MCQA contains 100 (Few Extra Labeled) or no (No Extra Labeled) labeled samples. UDIT is applied only to EXQA, CBQA, and MCQA.}

\begin{table*}[t]
\centering
\small
\begin{tabular}{lrcc|ccc|ccc|c|c}
\toprule
\multirow{2}{*}{Method} & \multirow{2}{*}{Size} & \multicolumn{2}{c|}{Coref.} & \multicolumn{3}{c|}{NLI} & \multicolumn{3}{c|}{Sentence Comp.} & WSD  & \multirow{2}{*}{Avg.} \\ \cmidrule(l){3-11}
                        &                       & WSC          & Wino.        & CB     & RTE    & ANLI   & COPA     & H-Swag     & Story    & WiC  &                       \\ \midrule
Vannila-IT              & 700M                  & 53.0         & 53.4         & 53.0   & 69.2   & \textbf{34.8}   & 75.2     & 27.4       & 89.6     & 50.7 & 56.9                  \\
SelfSup-IT              & 700M                  & 49.2         & \textbf{56.4}         & 56.5   & 67.7   & 34.1   & 75.7     & 26.1       & 90.4     & 51.0 & 56.3                   \\ \midrule
UDIT                    & 700M                  & \textbf{55.6}         & 53.5         & \textbf{57.8}   & \textbf{73.1}   & 33.5   & 74.8     & \textbf{30.1}       & 88.7     & 53.5 & \textbf{57.8}                  \\
UDIT + SelfSup-IT         & 700M                  & 53.4         & 55.7         & 54.2   & 65.5   & 33.6   & \textbf{76.7}     & 28.5       & \textbf{90.6}     & \textbf{56.6} & 57.2      \\ \midrule \midrule
Vannila-IT                 & 3B                    & 61.2         & 52.5         & 42.6   & 73.7   & 36.3   & 79.2     & 27.4       & 89.9     & 52.0 & 57.2                  \\
Vannila-IT                 & 11B                   & 61.4         & 59.9         & 70.1   & 80.8   & 43.6   & 90.0     & 33.6       & 92.4     & 56.6 & 65.4                  \\ \bottomrule
\end{tabular}
\caption{The scores of classification accuracy in the ``Full Labeled Data'' scenario. The \textbf{boldfaced} scores are the best results among the methods based on the 700M model. We reprint the results of the 11B model from \citet{t0} but re-evaluate the 3B model because the numerical values of this size are not provided in \citet{t0}.}
\label{tab:exp_full}
\vspace{-1em}
\end{table*}

By comparing the \uline{solid} and \dashuline{dashed} lines, we conclude that training on more tasks and instructions is beneficial, even if some tasks contain only 100 samples, which is consistent with Figure \ref{fig:setting}. Also, by comparing the \textcolor[RGB]{255,127,14}{orange} and \textcolor[RGB]{31,119,180}{blue} lines, we can see that UDIT leads to further improvement when applied to those data-scarce tasks, regardless of the existence of the 100 labeled samples. From Figure \ref{fig:task_cluster}, we notice the performance drop when adding TC, which matches the observation in \citet{zero-prompt} that not all the task clusters are helpful. But in general, the performance of IT has a positive correlation with the task number and diversity.

\subsubsection{Full Labeled Data}
\label{sec:full_labeled_data}
When the labeled data are sufficient, we can also mix them with the pseudo-labeled data to perform IT. Table \ref{tab:exp_full} shows that
adding 10K pseudo-labeled data can improve the IT performance, making our 700M model outperform the 3B model with Vanilla-IT. But increasing labeled data to 50k only leads to little further improvement (Figure \ref{fig:setting}). This indicates that the pseudo-labeled data do not merely contribute to the data amount per task. We conjecture that these samples also help avoid overfitting to the domain of specific datasets during IT, owing to the domain diversity of unlabeled corpora, which will be further analyzed in Section \ref{sec:exp_domain}.

\subsubsection{Language Generation Tasks}
\begin{table}[t]
\setlength\tabcolsep{4pt}
    \small
    \centering
    \begin{tabular}{llccc}
    \toprule
      Setting       & Method            & SQuAD & Story & Avg.\\ \midrule
\multirow{5}{*}{None} & DirectZS                 & \ \ 8.6      & \ \ 8.7       & \ \ 8.6\\ \cmidrule(lr){2-5}
                     & ExtraLM         &  13.3        &    15.2          & 14.2 \\
                     & SelfSup-IT         &  12.3        &   15.3      &  13.8  \\ \cmidrule(lr){2-5}
                     & UDIT               & 14.9     & 15.7      & 15.3 \\
                     & UDIT + SelfSup-IT        & \textbf{16.3}     & \textbf{17.2}       & \textbf{16.8} \\ \midrule \midrule
\multirow{4}{1.1cm}{Few\\Tasks} & Vanilla-IT & 25.9 & \textbf{16.3} & 21.1\\
                               & SelfSup-IT & 25.3 & 15.4 & 20.4\\ \cmidrule(lr){2-5}
                               & UDIT & 25.6 & 15.9 & 20.8 \\ 
                               & UDIT + SelfSup-IT & \textbf{27.3} & 15.2 & \textbf{21.2}\\ \midrule
\multirow{4}{1.1cm}{Few\\Datasets}  & Vanilla-IT & 17.5 & 14.5 & 16.0 \\
                               & SelfSup-IT & 19.5 & 15.4 & 17.4 \\ \cmidrule(lr){2-5}
                               & UDIT & \textbf{27.0} & 16.4 & \textbf{21.7} \\
                               & UDIT + SelfSup-IT & 25.5 & \textbf{16.7} & 21.1 \\ \midrule
\multirow{6}{1.1cm}{Few\\Samples}  & Vanilla-IT & 22.9 & 15.3 & 19.1 \\
                               & SelfSup-IT & 25.1 & 15.9 & 20.5\\
                               & DataAug-IT & 24.0 & 14.4 & 19.2 \\ \cmidrule(lr){2-5}
                               & UDIT & 25.4 & 15.4 & 20.4 \\ 
                               & UDIT + SelfSup-IT & \textbf{25.6} & \textbf{16.0} & \textbf{20.8}\\ 
                               & UDIT + DataAug-IT & 24.4 & 14.7 & 19.5\\\midrule \midrule
\multirow{3}{*}{Full}   & Vanilla-IT   & 28.5    & 15.3 & 21.9 \\
                        & SelfSup-IT & \textbf{30.3} & 15.2 & 22.8 \\ \cmidrule(lr){2-5}
                        & UDIT & 28.6 & \textbf{19.2} & \textbf{23.9} \\ 
                        & UDIT + SelfSup-IT & 29.4 & 17.5 & 23.4 \\ \bottomrule
                        
    \end{tabular}
    \caption{Results on language generation tasks. We report the average Rouge-L score across different testing instructions. ``None'' represents the ``No Labeled Data'' scenario. ``Few Tasks/Datasets/Samples'' means the three settings in ``Few Labeled Data'' discussed in Section \ref{sec:little_labeled_data}. ``Full'' represents ``Full Labeled Data''. All experiments are based on the 700M model.}
    \label{tab:exp_gen}
    \vspace{-0.5em}
\end{table}

We also test the instruction-tuned models on two language generation tasks. From Table \ref{tab:exp_gen}, we observe the similar phenomenon that UDIT improves IT the most in all scenarios. We also notice that self-supervised training is more beneficial to language generation than classification. This is likely because the self-supervised tasks include Next Sentence Generation and Next Phrase Generation~\cite{self-sup}, which resemble the generation tasks used in the zero-shot evaluation.

\subsection{Discussion}
Based on the results in Section \ref{sec:res}, we conclude that UDIT is effective under all three settings on both classification and generation tasks. 

We observe that UDIT brings larger improvements to Natural Language Inference (NLI) and Sentence Completion (Sentence Comp.) compared to Coreference Resolution (Coref.) and Word Sense Disambiguation (WSD), which resembles the phenomenon of Vanilla-IT. We suspect that our training tasks are mostly sentence-level, while the tasks in Coref. and WSD are word-level. Although IT enables cross-task generalization for PLMs, it is still challenging to generalize from sentence-level tasks to word-level tasks. This also emphasizes the importance of using instructions from more diverse tasks for IT. 

Besides, we also find that the performance variance across different testing instructions is high on some tasks (Figure \ref{fig:variance} in Appendix \ref{app:variance}), which is consistent with the observations in \citet{t0}. Reducing the sensitivity of PLMs to prompts and instructions has been largely discussed in previous literature~\cite{calibrate, prompt-consistency}. Most of these methods are applicable to our settings.

\section{Analysis}

\begin{table}[t]
\centering
\small
\begin{tabular}{lllcc}
\toprule
Method & Labeled & Pseudo & CLS & GEN \\ \midrule
No IT & -    & -        &   43.0       & \ \ 8.6        \\ \midrule
\multirow{2}{*}{Vannila-IT} 
      & Random & -      &   44.8       & \ \ 8.6       \\
      & Correct & -     &   56.9       & 15.3       \\ \midrule
\multirow{4}{*}{UDIT}
      & -    & Random   &   43.9       & 13.8       \\
      & -    & Correct  &   49.6       & 21.9       \\
      & Correct & Random &  52.3       & 22.1         \\
      & Correct & Correct & \textbf{57.8}       & \textbf{23.9}      \\ \bottomrule
\end{tabular}
\caption{The results on comparing whether we assign random labels to labeled and pseudo-labeled data. ``-'' means that we do not use the corresponding data. ``CLS'' and ``GEN'' stand for the average performance on the classification and generation tasks, respectively.}
\label{tab:exp_rand}
\end{table}

\subsection{Effect of Instruction Tuning}
IT brings two effects: (1) helping the model get familiar with the input form containing human instructions and (2) enabling the model to learn the mapping between the instructions and task semantics. To differentiate these effects, we construct tasks that do not match the instructions by randomly setting the labels in the labeled and pseudo-labeled samples. As shown in Table \ref{tab:exp_rand}, although we randomize the labels, the results of IT are still slightly better than No IT, suggesting that the input form matters. Furthermore, a large performance gap exists between the random and correct labels, indicating that the model learns the instruction-task mapping in addition to the instruction form.

\begin{table}[t]
\centering
\small
\begin{tabular}{llcc}
\toprule
Setting               & Method           & CLS & GEN \\ \midrule
\multirow{3}{*}{None} & DirectZS         &   43.0         & \ \ 8.6       \\
                      & UDIT (Single)    &   48.3         & 14.6       \\
                      & UDIT (Multiple)  &  \textbf{49.6} & \textbf{15.3}  \\ \midrule
\multirow{3}{*}{Full} & Vanilla-IT               &   56.9         & 21.9           \\
                      & UDIT (Single)    &   57.0         & 21.1           \\
                      & UDIT (Multiple)  &  \textbf{57.8} & \textbf{23.9}  \\ \bottomrule
\end{tabular}
\caption{The results when we use the pseudo-labeled data of single or multiple domains
for UDIT. ``CLS'' and ``GEN'' stand for the average performance on classification and generation tasks, respectively. ``None/Full'' represents the ``No/Full Labeled Data'' scenario.}
\label{tab:exp_domain}
\end{table}

\subsection{Effect of Domain Diversity}
\label{sec:exp_domain}
As described in Section \ref{sec:overview}, our unlabeled data are a mixture of multi-domain plain-text corpora. To investigate the domain diversity effect, we construct pseudo-labeled data only from Wikipedia for the task clusters other than SENT and TC, where we still use IMDB Review and CC-News. This ensures that each cluster contains a single domain. We also maintain the same amount of training samples as the multi-domain circumstance. From the results in Table \ref{tab:exp_domain}, we observe that reducing the domain diversity hurts UDIT. In the ``No Labeled Data'' scenario, the performance of UDIT mostly comes from the additional instructions. But in the ``Full Labeled Data'' scenario, the domain diversity contributes most to the improvement of UDIT.

\begin{figure}[t]
    \subfigbottomskip=0cm
    \subfigure[Data Amount] { \label{fig:data_quantity} 
    \centering
    \includegraphics[width=0.49\linewidth]{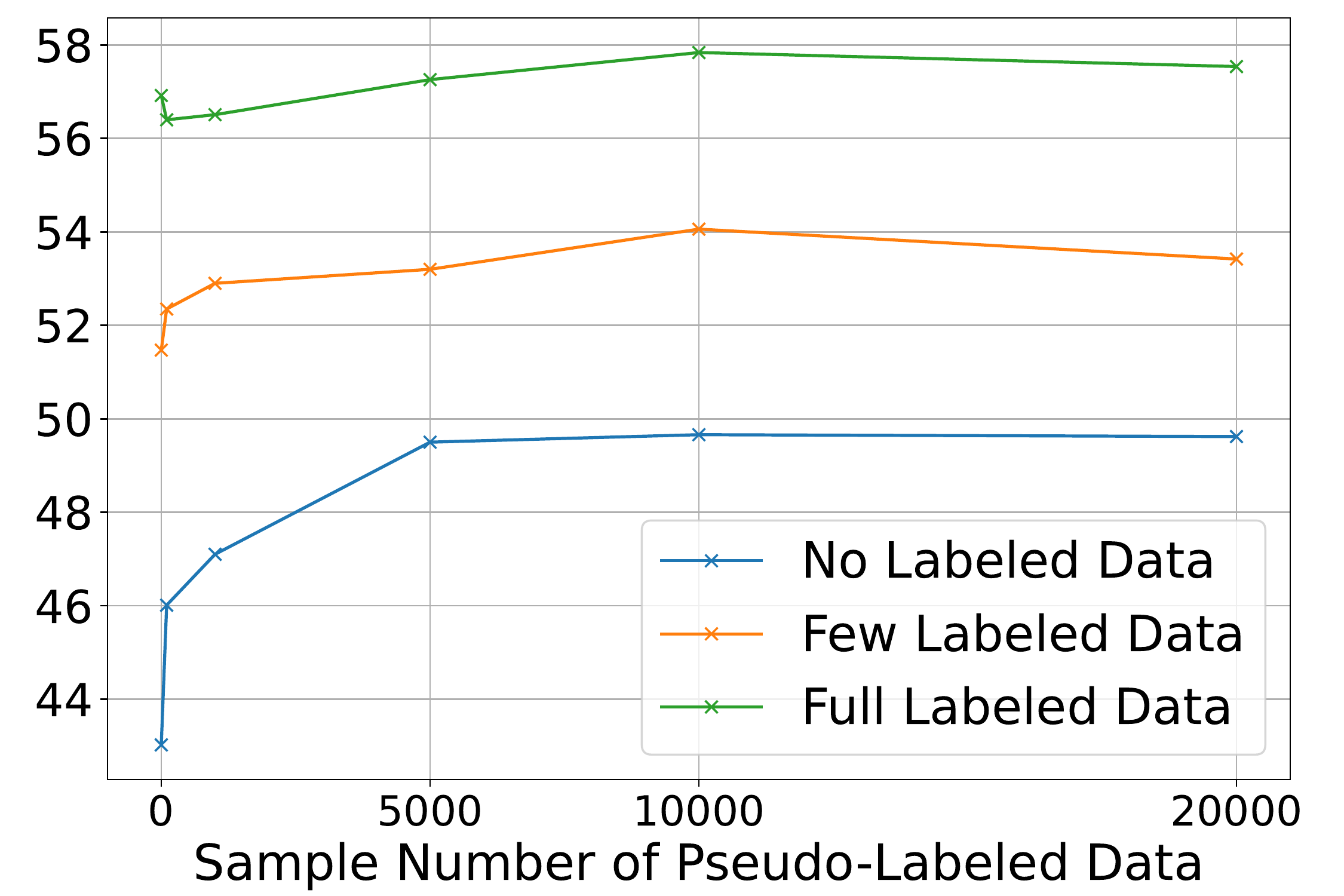} 
    }
    \subfigure[Individual Task Clusters] { \label{fig:data_quality} 
    \centering
    \includegraphics[width=0.45\linewidth]{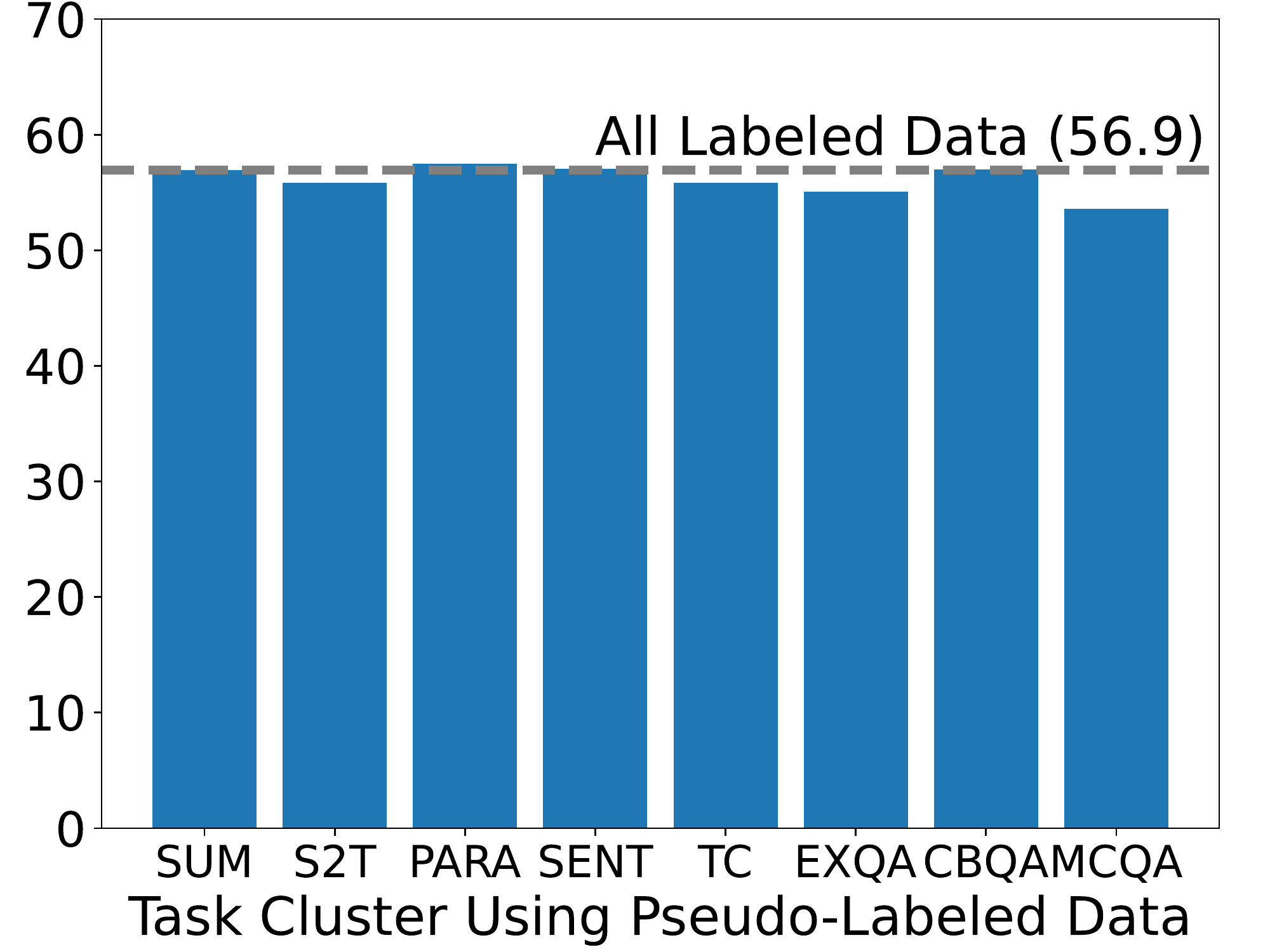} 
    } 
    \caption{The effect of the data amount and individual task clusters. The y-axis means the average results on the classification tasks.} 
    \label{fig:data_qq} 
    \vspace{-0.5em}
\end{figure} 

\subsection{Effect of Data Amount}
\label{sec:data_amount}


Since the pseudo-labeled data are constructed from the plain-text corpus, we can obtain numerous training samples for UDIT. However, as shown in Figure \ref{fig:data_quantity}, the performance converges when the number of pseudo-labeled training samples per task reaches 10k in all the three scenarios we consider in Section \ref{sec:res}. This is different from other methods using unlabeled data, such as self-supervised pre-training, where increasing the data amount continuously improves downstream performance~\cite{roberta,scaling_law}. These results suggest that UDIT is not data-hungry and does not consume much more training resources than Vanilla-IT.


\subsection{Effect of Individual Task Clusters}
The pseudo-labeled data inevitably contain noises which may hurt the model performance. Therefore, we investigate the influence of these noises in each task cluster. We choose one task cluster at a time, replace the labeled samples with pseudo-labeled samples, and perform IT on the mixed data. In Figure \ref{fig:data_quality}, we can see that using pseudo-labeled data in MCQA affects the zero-shot performance the most. But in general, replacing one task cluster does not bring much influence. This means that pseudo-labeled data in each cluster are of high quality and UDIT is robust to the noises in individual task clusters. However, comparing Table \ref{tab:exp_none} and Table \ref{tab:exp_full}, we find that UDIT still has a performance drop that cannot be ignored when replacing all the labeled data. It is probably caused by the noise accumulation in the pseudo-labeled data of multiple tasks. We leave how to further reduce the noises in the pseudo-labeled data as future work.


\section{Conclusion and Future Work}

In this work, we investigate performing IT with unlabeled data for zero-shot cross-task generalization. We first empirically find that the IT performance is largely restricted by the number of distinct tasks, instructions, and training samples in data-scarce tasks. Then, we propose UDIT to take better advantage of the instructions by constructing pseudo-labeled data from the unlabeled plain texts. Through UDIT, it is possible to perform IT with unlabeled data when there are few or no human-annotated samples, which offers a better way to incorporate unlabeled data compared with other approaches. Through comprehensive analysis, we find that the domain diversity and the matching between the pseudo-labeled data and corresponding instructions are essential for UDIT. In contrast, noises in individual task clusters and colossal data amount are less influential. There are three directions for future work: (1) Designing automatic and generalizable methods to construct pseudo-labeled data for instruction tuning. (2) Mining novel instructions from the unlabeled corpus to enlarge the amount of instructions during training. (3) Further denoising the pseudo-labeled data built from unlabeled plain texts.

\section*{Limitations}
The limitation of our work is that the process of constructing pseudo-labeled data from unlabeled plain texts still needs manual design. Although the strategies we use are easy to implement and our pseudo-labeled data have covered a big part of classic NLP tasks, there may exist some ``hard tasks'' where finding suitable methods to construct high-quality pseudo-labeled data is not easy. However, this is not a severe problem in practice because UDIT boosts instruction learning for zero-shot cross-task generalization. This means we can still improve the performance on the ``hard tasks'' with UDIT based on the pseudo-labeled data from the ``easy tasks''. We believe that more generalizable and elaborate data construction methods would further improve performance. We leave this as future work, and the findings in this work can guide the design of these methods.


\section*{Acknowledgements}
This paper was supported by the National Key Research and Development Program of China (No. 2021ZD0113304), the National Science Foundation for Distinguished Young Scholars (with No. 62125604), and
the NSFC projects (Key project with No. 61936010 and
regular project with No. 61876096). This work was also
supported by the Guoqiang Institute of Tsinghua University, with Grant No. 2019GQG1 and 2020GQG0005, and
sponsored by Tsinghua-Toyota Joint Research Fund.

\bibliography{anthology,custom}
\bibliographystyle{acl_natbib}

\clearpage

\appendix

\section*{Appendices}
\section{Data Information}
\label{app:data_info}
\subsection{Training Tasks}
Following \citet{t0}, we adopt 8 task clusters containing 36 datasets. The datasets and the number of instructions in each cluster are shown in Table \ref{tab:app_data_train}. All instructions are taken from the Public Pool of Prompts (P3)~\cite{p3}\footnote{\url{https://github.com/bigscience-workshop/promptsource}}.

\subsection{Evaluation Tasks}
We evaluate our model on 4 text classification task clusters and 2 language generation task clusters. The text classification task clusters and datasets include: (1) \textit{Coreference Resolution} (Coref.): WSC and Winogrande (Wino.)~\cite{winogrande}; (2) \textit{Natural Language Inference} (NLI): CB~\cite{cb}, RTE~\cite{rte}, and ANLI-R1~\cite{anli}; (3) \textit{Sentence Completion} (Sentence Comp.): COPA~\cite{copa}, HellaSwag (H-Swag)~\cite{hella_swag}, and Story Cloze~\cite{roc_story}; (4) \textit{Word Sense Diasambiguation} (WSD): WIC~\cite{wic}. The language generation task clusters and datasets include: (5) \textit{Question Generation} (QG): SQuAD~\cite{squad}; (6) \textit{Open-Ended Natural Language Generation} (ONLG): Roc Story~\cite{roc_story}. All instructions are obtained from the Public Pool of Prompts (P3)~\cite{p3}.

\subsection{Unlabeled Data}
Our unlabeled plain texts consist of the multi-domain corpus, including BookCorpus~\cite{bookcorpus} (5.5G), Wikipedia~\cite{wikidump} (20G), CC-News~\cite{cc_news} (1.7G), OpenWebText~\cite{openwebtext} (10G), IMDB Review~\cite{imdb} (65M). We access these data from the HuggingFace Datasets~\cite{datasets}\footnote{\url{https://huggingface.co/datasets}}. For OpenWebText, we randomly sample 10GB sample from the original 38GB samples to balance the data sources.

\begin{table}[t]
    \centering
    \small
    \begin{tabular}{l|r|p{4.3cm}}
    \toprule
     Cluster & \#Instr. & Datasets \\ \midrule
     MCQA    &  80      & COS-E~\cite{cos-e}, DREAM~\cite{dream}, QuAIL~\cite{quail}, QuaRTz~\cite{quartz}, Social-IQA~\cite{social_i_qa}, WiQA~\cite{wiqa}, CosmosQA~\cite{cosmos_qa}, QASC~\cite{qasc}, QUAREL~\cite{quarel}, SciQ~\cite{sciq}, Wiki-Hop~\cite{wikihop}\\ \midrule
     EXQA    &  46      & Adversarial-QA~\cite{adversarial_qa}, Quoref~\cite{quoref}, ROPES~\cite{ropes}, DuoRC~\cite{duorc}\\ \midrule
     CBQA    &  26      & Hotpot-QA~\cite{hotpotqa}, Wiki-QA~\cite{wikiqa}\\ \midrule
     SENT    &  43      & Amazon~\cite{amazon}, App-Reviews~\cite{app_reviews}, IMDB~\cite{imdb}, Rotten-Tomatoes~\cite{rotten_tomatoes}, Yelp~\cite{yelp}\\ \midrule
     TC      &  29      & AG-News~\cite{yelp}, DBPedia~\cite{dbpedia}, TREC~\cite{trec}\\ \midrule
     S2T     &  14      & Common-Gen~\cite{commongen}, Wiki-Bio~\cite{wiki_bio}\\ \midrule
     SUM     &  41      & CNN-Daily-Mail~\cite{cnn_dm}, Gigaword~\cite{gigaword}, MultiNews~\cite{multi_news}, SAMSum~\cite{samsum}, XSum~\cite{xsum}\\ \midrule
     PARA    &  25      & MRPC~\cite{mrpc}, PAWS~\cite{paws}, QQP~\cite{qqp}\\
     \bottomrule
    \end{tabular}
    \caption{Task clusters and datasets use for training. ``\#Instr.'' means the number of instructions in each cluster.}
    \label{tab:app_data_train}
\end{table}

\begin{table}[t]
    \centering
    \small
    \begin{tabular}{l|l|r|r}
    \toprule
    Cluster & Dataset   & \#Valid Set & \#Instr.  \\
    \midrule
    \multirow{2}{1cm}{Coref.}
            & WSC       &  104      &   10      \\
            & Winogrande & 1,267    &    5      \\ \midrule
    \multirow{3}{1cm}{NLI}
            & CB        &  57       &   15      \\
            & RTE       &  277      &   10      \\
            & ANLI-R1   &  1,000     &   15      \\ \midrule
    \multirow{3}{1cm}{Story\\ Comp.}
            & COPA      &  100      &   12      \\
            & HellaSwag &  10,042   &   5       \\
            & Story Cloze & 1,871   &   5       \\ \midrule
    \multirow{1}{1cm}{WSD}
            &  WIC      &  637      &   10      \\ \midrule
    \multirow{1}{1cm}{QG}
            &  SQuAD    &  10,570   &   2       \\ \midrule
    \multirow{1}{1cm}{ONLG}
            & Roc-Story &  1,871    &   1       \\
    \bottomrule
            
    \end{tabular}
    \caption{Task clusters and datasets for zero-shot cross-task evaluation. ``\#Valid Set'' stands for the number of samples in the validation split of each dataset, which we use as the test sets. ``\#Instr''. means the number of instructions in each dataset.}
    \label{tab:app_data_test}
\end{table}

\section{More Training Details}
\label{app:hyper_param}
We run IT on a 700M T5 model\footnote{\url{https://huggingface.co/liangtaiwan/t5-v1_1-lm100k-large}}. The max input sequence lengths of the encoder and the decoder are 512 and 128, respectively. We first run Vanilla-IT to select the hyper-parameters that yield the best performance on the validation splits of the training datasets. Then, we fix the hyper-parameters in all our experiments. We search for the learning rate in [3e-5, 5e-5, 1e-4], the batch size in [512, 1024, 2048], and the max training steps in [10K, 30K]. We finally set the learning rate to 5e-5, batch size to 1024, and the max training steps to 10K for both Vanilla-IT and UDIT. We use the Adam optimizer~\cite{adam} with $\beta_1=0.9$, $\beta_2=0.999$, $\epsilon=1e-8$, and $\text{weight\_decay}=0.01$. We follow \citet{flan} to balance each dataset by treating each task at most 3K samples per instruction for sampling. 

To improve the training efficiency, we adopt the mixed-precision training~\cite{mixed-fp} and ZeRO (stage-1)~\cite{zero} implemented in DeepSpeed~\cite{deepspeed}\footnote{\url{https://github.com/microsoft/DeepSpeed}}. Note that the T0-3B model is evaluated in FP32 precision. Our experiments are all conducted on the NVIDIA 32G V100 GPU. We use two GPUs for each run of IT, which completes in about 12 hours, depending on the total training data amount. The inference of a single model occupies one GPU and takes about 10 minutes.

\section{More results}
\subsection{Other Choices of the ``Few Tasks'' Setting}
\label{app:few_task}
In Table \ref{tab:more_few_task}, we present the results when we use different task clusters as the data-sufficient cluster, as a complementary to Table \ref{tab:exp_little}. From the results, we can see that UDIT improves Vanilla IT in most cases. One exception is the language generation tasks when we train the model on EXQA. We think the reason is that some tasks in EXQA are also formulated to question generation tasks, which are too similar to the evaluation task and cover the effect of UDIT. Note that the zero-shot performance of Vanilla-IT on language generation tasks is really poor when the model is trained only on SENT, TC, or PARA, which mainly consist of text classification tasks. We observe that all output texts are biased to the labels of corresponding training datasets, which means the model overfits the text classification tasks during IT and fails to learn to follow instructions in unseen tasks.

\begin{table}[t]
    \centering
    \small
    \begin{tabular}{lcccc}
    \toprule
    \multirowcell{2}{Full-Data \\ Cluster}
             & \multicolumn{2}{c}{Classification}   & \multicolumn{2}{c}{Generation}  \\ \cmidrule(lr){2-5}
             &  Vanilla-IT         &      UDIT          & Vanilla-IT       &  UDIT          \\ \midrule
      MCQA   & 52.9         & \textbf{56.5} &      19.8         &  \textbf{21.1}     \\
      EXQA   & 47.9         & \textbf{51.3} &  \textbf{21.1}    &  20.8     \\
      CBQA   & 45.4         & \textbf{50.7} &     10.2          &  \textbf{17.5}     \\
      SENT   & 47.2         & \textbf{50.6} &    \ \ 0.2         &  \textbf{15.9}     \\
      TC     & 44.3         & \textbf{49.7} &    \ \ 1.1         &  \textbf{16.4}     \\
      S2T    & 42.6         & \textbf{50.8} &       14.0        &  \textbf{19.0}     \\
      SUM    & 42.7         & \textbf{50.4} &       16.1        &  \textbf{17.2}     \\
      PARA   & 46.4         & \textbf{49.4} &    \ \ 0.8         &  \textbf{15.9}     \\
      \bottomrule
    \end{tabular}
    \caption{The results when we choose different task clusters as the data-sufficient cluster in the ``Few Tasks'' setting. ``Full-Data Cluster'' means the data-sufficient task cluster. We report the average performance on the text classification and generation tasks.}
    \label{tab:more_few_task}
\end{table}

\subsection{Median Results Across Different Testing Instructions}
\label{app:res_median}
Following ~\citet{t0}, we also report the median of the  performances across different testing instructions in Table \ref{tab:exp_none_median}, Table \ref{tab:exp_little_median}, and Table \ref{tab:exp_full_median} as a supplement to the mean of the performances in Table \ref{tab:exp_none}, Table \ref{tab:exp_little}, and Table \ref{tab:exp_full}, respectively. Comparing different approaches, we can draw similar conclusions as Section \ref{sec:res} that UDIT offers a significantly better way to incorporate unlabeled data into IT and improves the zero-shot cross-task generalization. We also observe that the mean and median do not differ much on most datasets, except for CB where the median is much better. 

\begin{table*}[t]
\centering
\small
\begin{tabular}{lrcc|ccc|ccc|c|c}
\toprule
\multirow{2}{*}{Method} & \multirow{2}{*}{Size} & \multicolumn{2}{c|}{Coref.} & \multicolumn{3}{c|}{NLI} & \multicolumn{3}{c|}{Sentence Comp.} & WSD  & \multirow{2}{*}{Avg.} \\ \cmidrule(l){3-11}
                        &                       & WSC          & Wino.        & CB     & RTE    & ANLI   & COPA     & H-Swag     & Story    & WiC  &                       \\ \midrule
DirectZS                & 700M                  & 50.0         & 51.1         & 33.3   & 47.4   & 33.1   & 42.7     & 26.6       & 53.0     & 50.3 & 43.0                  \\
DirectZS                & 3B                    & 50.0         & 51.0         & 33.3   & 47.6   & 32.7   & 45.8     & 25.6       & 55.6     & 50.6 & 43.6                  \\
DirectZS                & 11B                   & \textbf{57.7}& 50.7         & 33.9   & 51.8   & 32.7 & 55.0 & 27.7 & 48.8     & 50.3 & 45.4                  \\ \midrule
ExtraLM                 & 700M                  & 53.1         & 51.5         & 27.1   & 52.6   & 32.6   & 49.5     & 24.2       & 51.8     & 50.6 & 43.6                  \\
SelfSup-IT              & 700M                  & 50.8          &\textbf{54.6} & 41.7   & 52.9   & \textbf{33.2} & 51.0 & 23.7 & 50.7     & 50.2 & 45.4                  \\ \midrule
UDIT                    & 700M              &\underline{53.3}  & 52.6    & \textbf{50.0} & \underline{53.7} & \textbf{33.2} & \underline{57.8} & \underline{29.8} & \underline{68.0} & \underline{52.7} & \underline{50.0}   \\
UDIT + SelfSup-IT        & 700M               & \underline{53.3} & \underline{52.9} & \underline{45.8}   & \textbf{56.4} & \underline{32.8}   &  \textbf{58.8} & \textbf{29.9} & \textbf{71.5} & \textbf{52.8} & \textbf{50.5}                      \\ \bottomrule
\end{tabular}
\caption{The median classification accuracy of the experiments in Table \ref{tab:exp_none} (No Labeled Data).}
\label{tab:exp_none_median}
\end{table*}

\begin{table*}[t]
\setlength\tabcolsep{5.8pt}
\centering
\small
\begin{tabular}{llcc|ccc|ccc|c|c}
\toprule
 \multirow{2}{*}{Setting} & \multirow{2}{*}{Method} & \multicolumn{2}{c|}{Coref.} & \multicolumn{3}{c|}{NLI} & \multicolumn{3}{c|}{Sentence Comp.}   & WSD  & \multirow{2}{*}{Avg.} \\ \cmidrule(l){3-11}
                          &                         & WSC          & Wino.        & CB     & RTE    & ANLI   & COPA     & H-Swag             & Story & WiC  &                       \\ \midrule
\multirow{4}{1.1cm}{Few\\Tasks}
                          & Vannila-IT              & 50.8         & 54.4         & 62.5   & 53.7   & 31.2   & \textbf{61.4} & 28.2 & 50.7  & 50.8 & 49.2                  \\
                          & SelfSup-IT              & 50.8         & 54.9         & 51.4   & 52.9   & \textbf{33.3} & 58.0 & 27.4            & 53.0  & 50.3 & 48.0                  \\ \cmidrule(lr){2-12}
                          & UDIT                    & \textbf{53.8}& 55.0         & \textbf{68.8} & \textbf{57.3} & 32.4 & 58.3 & \textbf{29.4} & 66.8 & \textbf{54.9} & \textbf{53.0}            \\
                          & UDIT + SelfSup-IT        & 51.6         & \textbf{56.6} & 45.8  & 55.9   & \textbf{33.3}   & 59.9     & 26.7      & \textbf{67.9} & 51.9 & 50.0          \\ \midrule \midrule
\multirow{4}{1.1cm}{Few\\Datasets}
                          & Vannila-IT              & 50.0         & 52.2         & 39.6   & 49.8   & \textbf{33.0}   & 57.8     & 26.2       & 54.2     & 51.0 & 45.1                  \\
                          & SelfSup-IT              & 50.8         & 54.3         & 41.7   & 53.7   & \textbf{33.0}   & 56.2     & 27.3       & 60.2     & 50.4 & 47.5                   \\ \cmidrule(lr){2-12}
                          & UDIT                    & 50.8         & 50.9        & \textbf{56.2} & \textbf{66.9} & 30.7 & \textbf{68.2} & \textbf{30.2} & 71.3 & 51.5 & \textbf{53.0}            \\
                          & UDIT + SelfSup-IT        & \textbf{51.6} & \textbf{55.6} & 45.8 & 58.4 & 32.2 & 61.4 & 27.6      & \textbf{77.0}  & \textbf{52.6} & 51.4                    \\ \midrule \midrule
\multirow{6}{1.1cm}{Few\\Samples}
                          & Vanilla-IT              & 50.8         & 51.8         & 56.2   & 57.4   & 31.4   & 66.7     & 26.5              & 71.0     & 53.4 & 51.7                  \\
                          & SelfSup-IT              & 50.8         & 53.2         & 47.9   & 62.7   & 31.8   & 61.4     & 26.4              & 77.2     & 54.6 & 51.8                  \\
                          & DataAug-IT              & 50.8         & 52.2     & \textbf{60.6}& 54.6 & 30.8   & 64.8     & 28.0              & 72.0     & 53.0 & 51.9                  \\ \cmidrule(lr){2-12}
                          & UDIT                    & \textbf{53.1} & 52.7        & 56.2   & \textbf{65.1} & 31.4 & 69.8 & \textbf{29.6}     & 79.6  & 55.2 & \textbf{54.7}            \\
                          & UDIT + SelfSup-IT        & 52.3         & \textbf{54.3}& 52.1   & 55.7   & \textbf{32.7} & 70.3 & 28.9    & \textbf{82.0} & \textbf{56.0} & 53.6                     \\
                          & UDIT + DataAug-IT        & 50.8         & 51.3         & 56.2   & 61.0   & 31.8   & \textbf{72.9} & 29.1         & 80.2   & 52.6 & 54.2                      \\ \bottomrule
\end{tabular}
\caption{The median classification accuracy of the experiments in Table \ref{tab:exp_little} (Few Labeled Data).}
\label{tab:exp_little_median}
\end{table*}

\begin{table*}[t]
\centering
\small
\begin{tabular}{lrcc|ccc|ccc|c|c}
\toprule
\multirow{2}{*}{Method} & \multirow{2}{*}{Size} & \multicolumn{2}{c|}{Coref.} & \multicolumn{3}{c|}{NLI} & \multicolumn{3}{c|}{Sentence Comp.} & WSD  & \multirow{2}{*}{Avg.} \\ \cmidrule(l){3-11}
                        &                       & WSC          & Wino.        & CB     & RTE    & ANLI          & COPA     & H-Swag     & Story    & WiC  &                       \\ \midrule
Vannila-IT              & 700M                  & 52.3         & 53.1         & 62.5   & 68.8   & \textbf{35.0} & 76.3     & 27.5       & 89.8     & 50.7 & 58.0                  \\
SelfSup-IT              & 700M                  & 48.4         & \textbf{56.8}& 60.4   & 66.5   & 34.3          & 79.7     & 25.8       & 90.5     & 50.7 & 57.0                   \\ \midrule
UDIT                    & 700M                  & \textbf{53.9}& 53.4   & \textbf{70.8}& \textbf{73.2} & 33.3   & 78.6     & \textbf{29.0} & 88.3  & 53.3 & \textbf{59.3}                  \\
UDIT + SelfSup-IT         & 700M                  & 51.6         & 56.2         & 56.2   & 64.7   & 34.3          & \textbf{80.2} & 26.9  & \textbf{90.7} & \textbf{56.7} & 57.5      \\ \midrule \midrule
Vannila-IT              & 3B                    & 62.5         & 52.4         & 50.0   & 72.5   & 35.2          & 78.1     & 28.3       & 88.9     & 52.4 & 57.8                  \\
Vannila-IT              & 11B                   & 64.4         & 60.5         & 78.6   & 81.2   & 44.7          & 90.8     & 33.6       & 94.7     & 57.2 & 67.3                  \\ \bottomrule
\end{tabular}
\caption{The median classification accuracy of the experiments in Table \ref{tab:exp_full} (Full Labeled Data).}
\label{tab:exp_full_median}
\vspace{-1em}
\end{table*}

\subsection{Variance Across Instructions}
\label{app:variance}

\begin{figure*}[t]
    \centering
    \includegraphics[width=\linewidth]{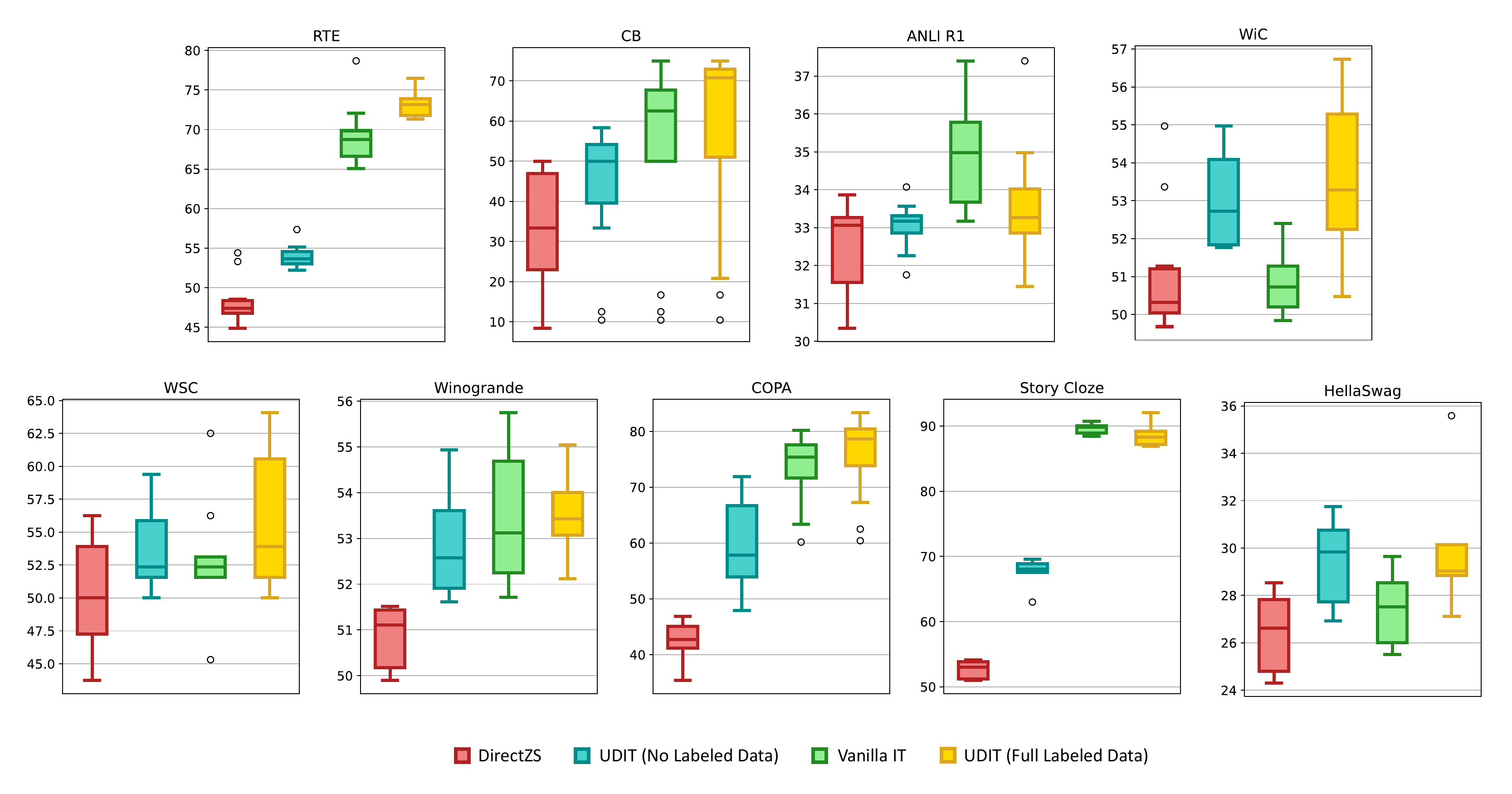}
    \caption{Evaluation results on classification tasks with variance across different instructions.}
    \label{fig:variance}
\end{figure*}

We draw the box plot of UDIT and some baselines under the ``No Labeled Data'' (Section \ref{sec:no_label_data}) and the ``Full Labeled Data'' (Section \ref{sec:full_labeled_data}) settings to show the variance across different instructions. From Figure \ref{fig:variance}, we can see that the results vary across instructions in all methods, which is also observed in \citet{t0}. There are plenty of studies on how to reduce the variance across prompts or instructions~\cite{prompt-consistency,calibrate,prompt_order}. Most of them can be combined with our methods.

\subsection{Human Evaluation on Pseudo-Labeled Data}
We conduct human evaluation on the pseudo-labeled data, and the results are shown in Table \ref{tab:human_eval}. For each task cluster, we randomly select 50 sample-instruction pairs and recruit 3 different annotators from Amazon Mechanical Turk\footnote{\url{https://www.mturk.com/}} to evaluate whether the pseudo-labeled sample is aligned with the instruction (scored as 1) or not (scored as 0). The final score for each task cluster is averaged over all the samples and 3 different annotators. From the results, we can see that although most of the pseudo-labeled samples make sense to humans, there inevitably exist some mislabeled samples that may be harmful to the model. We leave how to further denoise the pseudo-labeled data to future work. 

\begin{table*}[t]
    \centering
    \small
    \begin{tabular}{ccccccccc}
    \toprule
    Data Cluster & MCQA & EXQA & CBQA &  TC  & SENT & S2T  & SUM  & PARA \\\midrule
    Score & 0.91 & 0.64 & 0.65 & 0.88 & 0.83 & 0.77 & 0.76 & 0.89 \\ \bottomrule
    \end{tabular}
    \caption{Human evaluation results of the pseudo-labeled data.}
    \label{tab:human_eval}
\end{table*}

\section{Examples of Pseudo-Labeled Data}
\label{app:case}
We list a few examples of the pseudo-labeled data for the 8 task clusters in Table \ref{tab:case}. In MCQA, EXQA, and CBQA, although the constructing process relies on some assumptions, the pseudo-labeled data reflect the task semantics well and thus match the meanings of the corresponding instructions. We notice some incoherence and typos in the pseudo-labeled data, but this does not affect the general meanings of the sentences. For TC, SENT, and SUM, we find the pseudo-labeled data to be of high quality. For S2T and PARA, we observe that the pseudo-labeled data is much easier than the labeled data. This may harm conventional supervised learning since these data can hurt the model's ability to solve hard samples. However, we argue that this issue is not that severe in instruction learning because ``learning to follow instructions'' only requires the correct mapping between the instructions and the task semantics, which is satisfied in the pseudo-labeled data, despite its simpleness.

\begin{table*}[t]
    \centering
    \small
    \begin{tabular}{l|p{6cm}|p{7.8cm}}
    \toprule
    Task & Plain Texts & Pseudo-Labeled Data \\ \midrule
    MCQA      & "Just open-source it" is not realistic. I've received a couple of questions about fred following the failure of ... The big one is: "\underline{why don't you just open-source it as-is?}". My answer is: it's impractical, and it wouldn't help anyone as-is. & \textbf{Passage}: "Just open-source it" is not realistic. I've received a couple of questions about fred following the failure of ... \newline \textbf{Question}: why don't you just open-source it as-is? \newline \textbf{Answer}: My answer is: it's impractical, and it ... \newline \textbf{Option A}: My answer is: it's impractical, and it ... \newline
    \textbf{Option B}: Both Harrison and Andrew laughed ... \\ \midrule
    EXQA      & It lies west of Liniewo, east of Kościerzyna, and south-west of the regional capital Gdanśk. ... South-east of Kościerzyna, and south-west of the regional capital Gdanśk, it lies approximately south of \underline{Liniewo}.&  \textbf{Passage}: It lies west of Liniewo, east of Kościerzyna, and south-west of the regional capital Gdanśk. ... \newline \textbf{Question}: Where south-east of Kościerzyna, and south-west of the regional capital Gdanśk, it lies approximately south of? \newline \textbf{Answer}: Liniewo \\ \midrule
    CBQA      & Psychroflexus planctonicus is a gram-negative bacteria which has been isolated from the Lake Xiaochaidan in \underline{the Qinghai Province} In China. & \textbf{Question}: Where in China, Psychroflexus planctonicus is a gram-negative bacteria which has been isolated from the Lake Xiaochaidan in? \newline \textbf{Answer}: the Qinghai Province \\ \midrule
    TC        & URL: https://www.mydailyregister.com/\underline{sports}/14 \newline 501/eagles-topple-trimble-8-5 \newline  The Eastern baseball team trailed 2-0, two innings into Monday night’s Tri-Valley Conference Hocking Division showdown with ... & \textbf{Passage}: The Eastern baseball team trailed 2-0, two innings into Monday night’s Tri-Valley Conference Hocking Division showdown with ... \newline \textbf{Label}: Sports\\ \midrule
    SENT      & \underline{AMAZING!} The staff was so \underline{friendly}, \underline{welcoming} and the food was \underline{superb}!  & \textbf{Passage}: AMAZING! The staff was so friendly, welcoming and the food was superb! \newline \textbf{Label}: Positive \\ \midrule
    S2T  & Other features will help make \underline{Henn-na} the \underline{most} futuristic low-cost \underline{hotel} in the \underline{industry}. & \textbf{Keyword}: most; industry; hotel; Henn-na \newline \textbf{Text}: Other features will help make Henn-na the most futuristic low-cost hotel in the industry.\\ \midrule
    SUM       & \underline{Title: }NVIDIA Deep Learning Platform Gives Enterprise Customers Instant Access to AI ... \newline \underline{Passage:} Baidu and NVIDIA are long-time partners in advancing the state of the art in AI ... & \textbf{Passage}: Baidu and NVIDIA are long-time partners in advancing the state of the art in AI... \newline \textbf{Summary}: NVIDIA Deep Learning Platform Gives Enterprise Customers Instant Access to AI... \\ \midrule
    PARA      & You're \underline{likely} vulnerable to \underline{online} attacks. & \textbf{Sent1}: You're likely vulnerable to online attacks. \newline \textbf{Sent2}: You are incredibly vulnerable to online pilots. \newline \textbf{Label}: not\\
    \bottomrule
    \end{tabular}
    \caption{Examples of the unlabeled plain texts and the corresponding pseudo-labeled data for each task cluster. The important parts in the unlabeled plain texts for data construction are \underline{underlined}.}
    \label{tab:case}
\end{table*}

\end{document}